\documentclass{article}

\usepackage{PRIMEarxiv}
\usepackage[utf8]{inputenc} 
\usepackage[T1]{fontenc}    
\usepackage{hyperref}       
\usepackage{url}            
\usepackage{booktabs}       
\usepackage{amsfonts}       
\usepackage{nicefrac}       
\usepackage{microtype}      
\usepackage{lipsum}
\usepackage{fancyhdr}       
\usepackage{graphicx}       
\usepackage{amsmath}        
\usepackage{longtable}      
\graphicspath{{media/}}     

\title{Research on Individual Trait Clustering and Development Pathway Adaptation Based on the K-means Algorithm
}

\author{
  Qianru Wei\textsuperscript{1,}\thanks{These authors contributed equally to this work.}, Jihaoyu Yang\textsuperscript{1, $\ast$}, Cheng Zhang\textsuperscript{1}, Jinming Yang\textsuperscript{2,}\thanks{Corresponding author.} \\
  \textsuperscript{1} Ulster College, Shaanxi University of Science \& Technology, Xi'an, China \\
  \textsuperscript{2} Complex Lab, School of Computer Science and Engineering, \\ University of Electronic Science and Technology of China, Chengdu, China \\
  \texttt{\{weiqianru, 202315030305, 4247\}@sust.edu.cn} \\
  \texttt{yangjinming@std.uestc.edu.cn}
}

\begin{document}
\nocite{ref1,ref2,ref3,ref4,ref5,ref6,ref7,ref8,ref9,ref10,ref11,ref12,ref13,ref14,ref15,ref16,ref17,ref18,ref19,ref20,ref21,ref22,ref23,ref24,ref25,ref26,ref27,ref28,ref29,ref30,ref31}
\maketitle

\begin{abstract}
With the development of information technology, the application of artificial intelligence and machine learning in the field of education shows great potential. This study aims to explore how to utilize K-means clustering algorithm to provide accurate career guidance for college students. Existing methods mostly focus on the prediction of career paths, but there are fewer studies on the fitness of students with different combinations of characteristics in specific career directions. In this study, we analyze the data of more than 3000 students on their CET-4 scores, GPA, personality traits and student cadre experiences, and use the K-means clustering algorithm to classify the students into four main groups. The K-means clustering algorithm groups students with similar characteristics into one group by minimizing the intra-cluster squared error, ensuring that the students within the same cluster are highly similar in their characteristics, and that differences between different clusters are maximized. Based on the clustering results, targeted career guidance suggestions are provided for each group. The results of the study show that students with different combinations of characteristics are suitable for different career directions, which provides a scientific basis for personalized career guidance and effectively enhances students' employment success rate. Future research can further improve the precision of clustering and the guidance effect by expanding the sample size, increasing the feature variables and considering external factors.
\end{abstract}


\section{Introduction}
With the development of information technology, the application of artificial intelligence and machine learning technologies in various fields has been deepening, especially in the field of education, which has shown great potential. At present, many researches have been conducted to explore how to utilize these advanced technologies to enhance students' employability. As the core content of student work, college students' employment guidance service is highly valued by the state and society, and its importance is self-evident. Through effective employment guidance, it can not only help students better understand themselves and plan their future development, but also provide scientific decision-making support for university administrators. Therefore, it is of great theoretical and practical significance to study how to utilize artificial intelligence technology for employment guidance.

Person-job matching is the core concept and important purpose of career planning. In terms of theories related to job matching, Frank Parsons created the trait factor theory, laying the foundation of career development theory. Subsequently, the American occupational psychologist Holland created the personality type theory, which further enriched the theoretical foundation of this field. These theories emphasize the importance of matching between individual traits and the occupational environment for career success. With the development of big data and machine learning techniques, researchers have begun to combine these theories with modern technology to explore more precise and personalized approaches to career guidance.

For example, the intelligent career guidance system developed by Vignesh et al. (2021) emphasized student skill assessment-based career Path Prediction \cite{ref14}. Railean's (2022) Init2Winit application developed through AI and machine learning techniques also significantly improved the accuracy and personalization of career advice. Although there have been a number of studies that have achieved remarkable results, there are still some problems and shortcomings. For example, most of the existing studies focus on the prediction of career paths, while there are fewer studies on the fitness of students with different combinations of characteristics in specific career directions. In addition, how to effectively integrate multiple student characteristics (e.g., academic performance, personality traits, and leadership experience) in practical applications to provide more accurate career guidance still needs to be further explored.

Based on this, this study adopts the K-means clustering algorithm to categorize students into different groups by analyzing multiple data such as students' CET-4 scores, GPA, personality traits, and experience as student leaders, and to provide targeted career guidance suggestions for each group. The main purpose of this study is to explore how to utilize K-means clustering algorithm to provide accurate career guidance for college students. Specifically, firstly, through data preprocessing and normalization, the feature data with different scales are converted into consistent measures; then, the K-means clustering algorithm is used to classify the students into four main groups; finally, based on the clustering results, specific career guidance suggestions are provided for each group.

The results of this study show that students with different combinations of features are suitable for different career directions, which provides a scientific basis for personalized career guidance and effectively improves the success rate of students' employment. Through this study, we not only verified the effectiveness of the K-means clustering algorithm in career guidance, but also provided a data-driven career guidance framework for universities. Future research can further improve the accuracy of clustering and the effectiveness of guidance by enlarging the sample size, increasing the feature variables, and considering external factors, so as to better serve the career development needs of students.

\section{Related Work}

\subsection{Career Guidance Evolution}
Several studies have addressed issues related to career guidance for university students. Pereira González et al. (2019) evaluated the career guidance needs of university students, revealing significant gaps in the support provided by faculty \cite{ref9}. Berkutova et al. (2022) proposed a systematic multi-level career guidance approach that integrates various educational stages, emphasizing early and continuous support \cite{ref3}. Huang (2022) explored integrating AI with civic education to establish an adaptive career counseling system, demonstrating improved recommendation accuracy and effectiveness \cite{ref6}. Wang (2019) discussed constructing an employment guidance system through school-enterprise cooperation, emphasizing the integration of external resources and practical experiences \cite{ref15}. Shestakova and Agaltinova (2022) analyzed the role of information technology in expanding career guidance opportunities, particularly through online platforms, which offer broader reach and interaction with potential applicants \cite{ref13}. Prokhorov and Pyadysheva (2020) described career guidance work in the "Advertising and Public Relations" program, emphasizing interactive methods to engage students \cite{ref10}. Fedulova et al. (2022) presented an innovative approach to career guidance through immersive educational experiences, enhancing students' understanding of professional activities.

\subsection{ML-Driven Decision Models}
These studies highlight the importance of holistic, technology-enhanced, and continuous career guidance to meet the evolving needs of university students, addressing issues such as inadequate faculty support, the need for multi-level guidance systems, and the integration of civic education and practical experiences. A significant body of research has focused on the application of machine learning for career guidance \cite{ref22}. For instance, Railean (2022) investigated the use of AI and machine learning to align career choices with educational attainment through the Init2Winit app, demonstrating the potential to reduce educational inequities \cite{ref11}. Vignesh et al. (2021) developed an intelligent career guidance system that predicts suitable career paths based on an objective assessment of students' skills, aiding in crucial decision-making stages \cite{ref14}. Similarly, Zahour et al. (2020) employed neural networks for the automatic classification of vocational guidance questions, aiding in the development of a new vocational guidance system, highlighting the efficacy of personalized career recommendations \cite{ref17}.

\subsection{K-means in Education}
The use of machine learning techniques, particularly K-means clustering \cite{ref20, ref21, ref24}, has been prevalent in various educational contexts \cite{ref25, ref26, ref27}. For example, García-Peñalvo et al. (2020) utilized hierarchical clustering and K-means to classify European countries based on desktop computer availability in schools, while Shen and Duan (2020) applied K-means for analyzing teaching satisfaction surveys, providing insights for improving teaching strategies \cite{ref12}. Handoko (2016) used K-means clustering to improve the quality of learning in higher education institutions, demonstrating its effectiveness in educational data mining \cite{ref5}. Yu et al. (2022) explored using K-means for university academic evaluation, showing significant improvements in accuracy by mapping educational data within Riemannian space \cite{ref16}. Similarly, Zhang et al. (2024) explored simulation-based machine learning models to predict student academic performance using big data, showcasing another application of data analytics in enhancing educational outcomes \cite{ref30}. Chau and Phung (2018) presented a transfer-learning-based clustering method optimized with kernel K-means for effective educational data clustering, addressing the challenges of data scarcity and enhancing clustering accuracy \cite{ref2}. Mammadova et al. (2020) utilized K-means for clustering educational content on social networks, personalizing learning materials and improving e-learning outcomes \cite{ref8}. These studies illustrate the versatility and effectiveness of machine learning and K-means clustering in educational contexts \cite{ref1, ref28, ref29}. However, they also reveal gaps such as the need for continuous support, integration of civic education, and dynamic adaptability to individual needs.

\section{Theory and calculation}
The theoretical foundation of this study lies in integrating classical person-job fit theory with modern unsupervised learning algorithms. Frank Parsons' trait-factor theory and Holland's personality type theory indicate that the key to career success lies in matching individual traits with occupational environments. However, traditional methods struggle to achieve this goal at scale and objectively. The K-means clustering algorithm offers a data-driven solution, whose core principle involves minimizing the sum of squared errors within clusters automatically partitioning students within high-dimensional feature spaces into internally homogeneous and externally heterogeneous groups, thereby transforming the abstract "person-job fit theory" into a computable and actionable mathematical model.

At the computational level, we first preprocess the collected raw data: CET-4 scores (320-550) and GPA (2.3-4.7) are normalized using the minimum-maximum normalization formula scaled uniformly to the [0,1] range to eliminate dimensional effects; simultaneously, personality traits (introversion/extroversion) and student leadership experience (yes/no) were encoded as 0 and 1 to enable numerical computation. Subsequently, the elbow rule was applied to determine the optimal cluster number $k=4$, identifying the inflection point where the rate of decrease in SSE begins to slow significantly within the curve showing SSE decline as $k$ increases.

Subsequently, the algorithm randomly initialized four centroids. Through iteration, each student was assigned to the nearest centroid, and the centroid positions were recalculated until convergence. Finally, we analyzed the centroid characteristics of each cluster (e.g., high GPA + high CET-4 + introverted personality) and mapped them to the most suitable career paths (e.g., technical roles), completing the practical transformation from theoretical model to personalized employment guidance recommendations.

\section{Methodology}

\subsection{Data collection}
We have collected data on nearly 5000 undergraduate graduates from S University in Western China. Excluding outliers in statistical data and students who were admitted or unemployed, there were over 3000 students who chose to work. For the purpose of statistics and calculations, we selected 3000 data points as the dataset. The dataset of this study covers a wide range of attributes such as students' academic performance, personality traits, and experience as student leaders.

The specific distribution of the collected data is as follows:
\begin{itemize}
    \item \textbf{CET-4 Score:} Reflects a student's level of English and ranges from 320 to 623.
    \item \textbf{GPA:} Reflects a student's academic performance, with scores ranging from 1.69 to 4.29.
    \item \textbf{Personality:} determined by the MBTI test and categorized as introvert (i) and extrovert (e).
    \item \textbf{Student\_Leader:} a two-valued variable indicating whether the student has served as a student leader (1 is yes, 0 is no).
    \item \textbf{Job types:} including management positions, sales positions, technical positions, product positions, other.
\end{itemize}

The sample size of the data was 3000 and the data was processed to meet the needs of the study. Here we show the data of 50 of these students in Table \ref{tab:students}.

\begin{longtable}{ccclcl}
\caption{Sample data of 50 students} \label{tab:students} \\
\toprule
\textbf{Serial Number} & \textbf{CET4} & \textbf{GPA} & \textbf{Personality} & \textbf{Student\_Leader} & \textbf{Job} \\
\midrule
\endfirsthead
\midrule
\textbf{Serial Number} & \textbf{CET4} & \textbf{GPA} & \textbf{Personality} & \textbf{Student\_Leader} & \textbf{Job} \\
\midrule
\endhead
\midrule
\multicolumn{6}{r}{\textit{Continued on next page}} \\
\endfoot
\bottomrule
\endlastfoot
1  & 409 & 4.51 & e & 1 & sales post \\
2  & 538 & 3.06 & e & 0 & product post \\
3  & 526 & 3.50 & e & 1 & management post \\
4  & 472 & 4.07 & i & 0 & technical post \\
5  & 453 & 3.72 & e & 1 & management post \\
6  & 532 & 4.34 & i & 0 & technical post \\
7  & 517 & 2.38 & e & 0 & sales post \\
8  & 327 & 4.15 & i & 0 & other \\
9  & 410 & 3.38 & e & 0 & product post \\
10 & 476 & 4.34 & e & 0 & product post \\
11 & 522 & 4.55 & e & 0 & sales post \\
12 & 452 & 2.57 & e & 0 & product post \\
13 & 501 & 2.30 & e & 1 & product post \\
14 & 483 & 3.79 & i & 0 & technical post \\
15 & 424 & 3.33 & e & 0 & product post \\
16 & 539 & 4.46 & e & 1 & product post \\
17 & 324 & 4.52 & i & 1 & other \\
18 & 425 & 4.70 & e & 1 & product post \\
19 & 403 & 3.68 & e & 0 & technical post \\
20 & 525 & 3.18 & e & 0 & product post \\
21 & 477 & 2.82 & e & 0 & product post \\
22 & 487 & 4.16 & e & 1 & product post \\
23 & 499 & 4.26 & i & 0 & sales post \\
24 & 518 & 3.61 & e & 1 & management post \\
25 & 422 & 2.60 & e & 1 & sales post \\
26 & 539 & 4.36 & e & 0 & sales post \\
27 & 385 & 3.07 & e & 0 & management post \\
28 & 427 & 3.21 & e & 1 & sales post \\
29 & 490 & 2.80 & e & 1 & management post \\
30 & 429 & 2.76 & e & 1 & product post \\
31 & 421 & 3.87 & e & 0 & technical post \\
32 & 516 & 4.59 & i & 0 & technical post \\
33 & 434 & 4.03 & e & 1 & technical post \\
34 & 530 & 3.88 & e & 1 & management post \\
35 & 466 & 3.77 & e & 1 & technical post \\
36 & 452 & 4.69 & i & 1 & other \\
37 & 548 & 3.95 & e & 1 & technical post \\
38 & 534 & 2.64 & e & 0 & product post \\
39 & 526 & 3.64 & e & 1 & product post \\
40 & 478 & 2.69 & e & 1 & product post \\
41 & 455 & 3.51 & e & 0 & product post \\
42 & 532 & 2.61 & e & 0 & product post \\
43 & 404 & 4.19 & e & 1 & product post \\
44 & 450 & 2.79 & e & 0 & sales post \\
45 & 410 & 4.00 & e & 0 & product post \\
46 & 515 & 3.99 & e & 0 & product post \\
47 & 422 & 4.56 & e & 1 & sales post \\
48 & 396 & 4.20 & e & 0 & sales post \\
49 & 503 & 3.34 & e & 1 & sales post \\
50 & 412 & 3.26 & e & 0 & product post \\
\end{longtable}

\begin{figure}[htbp]
  \centering
  \includegraphics[width=0.7\textwidth]{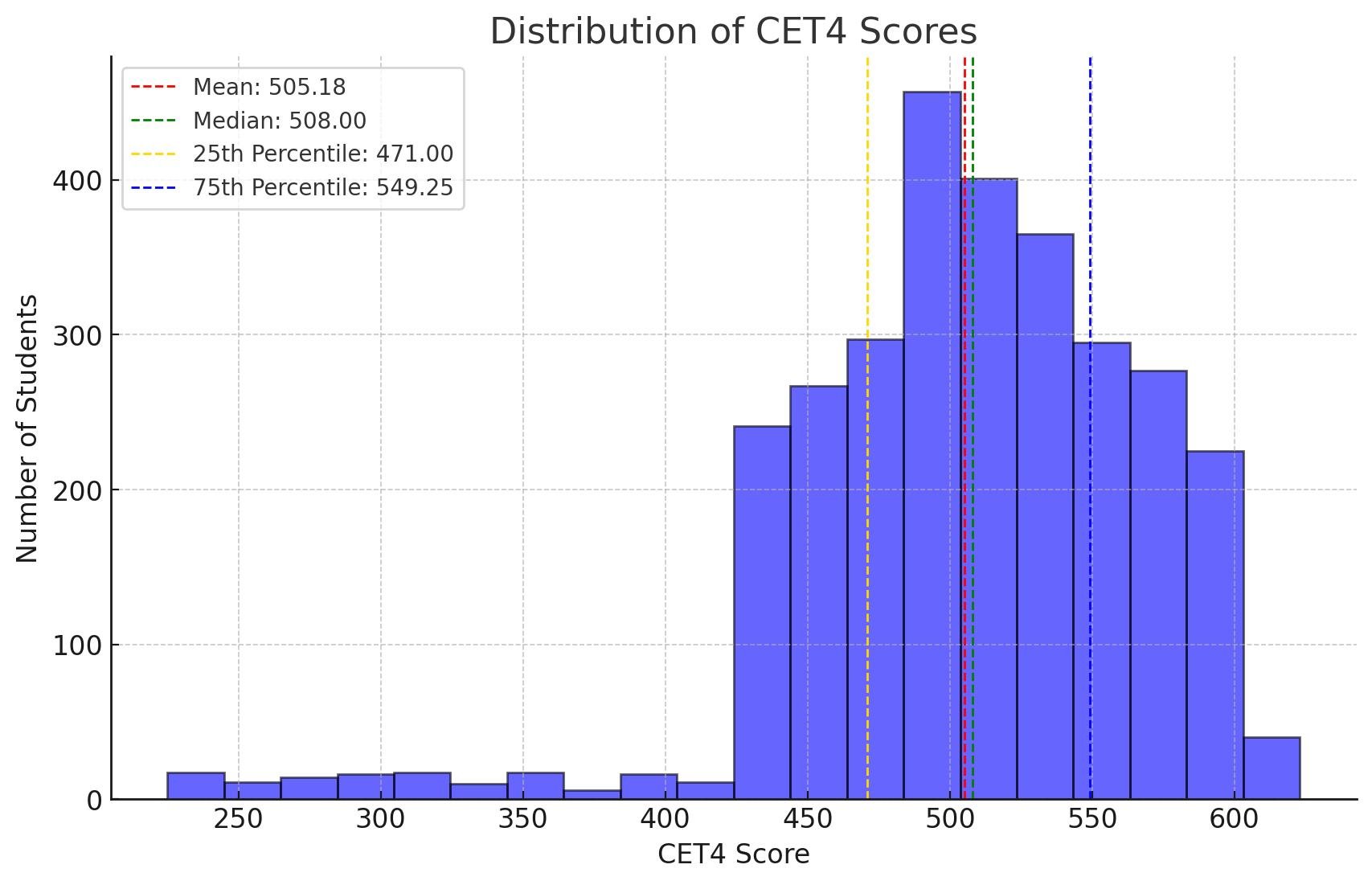}
  \caption{Histogram of students' CET-4 scores showing a left-skewed trend.}
  \label{fig:cet4}
\end{figure}

\begin{figure}[htbp]
  \centering
  \includegraphics[width=0.7\textwidth]{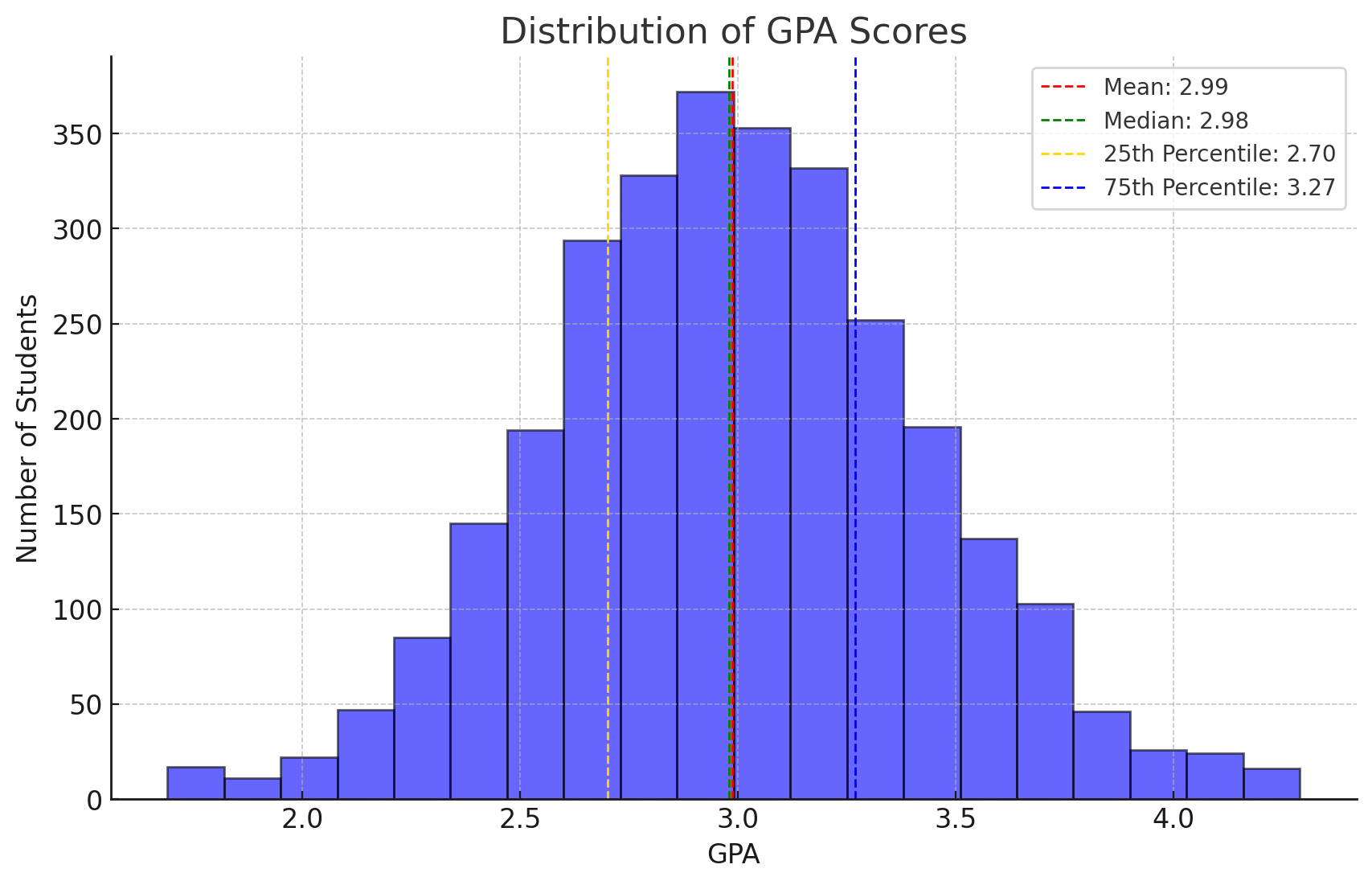}
  \caption{Histogram of students' GPA showing a roughly normal distribution.}
  \label{fig:gpa}
\end{figure}

The student's performance in terms of CET-4 score and GPA is shown in Figure \ref{fig:cet4} and Figure \ref{fig:gpa}.

The average score of CET-4 is 505.18, with a median of 508.00, indicating that the English proficiency of most students is relatively close to the average level. The 25th percentile is 471.00, indicating that these students may require more learning support and resources in English. The 75th percentile is 549.25, indicating strong English proficiency. Most students perform well in the CET-4 exam and have good English proficiency.

The average GPA is 2.99, with a median of 2.98, and most students' academic performance is concentrated around 3.0. The 25th percentile of GPA is 2.70, and these students need to put in more effort academically to improve their grades. Meanwhile, the 75th percentile is 3.27, indicating excellent academic performance among these students. The histogram shows that the distribution of GPA is roughly normal, reflecting the overall consistency and effectiveness of school education.

\subsection{Data Preprocessing}

\subsubsection{Data normalization processing}
In order to ensure the consistency of data processing and the fairness of the results, we normalized the CET-4 scores and GPAs in the dataset. The main purpose of normalization is to convert attributes of different scales to the same range of magnitude to eliminate the effect of numerical magnitude on the clustering results. The normalization formula is as follows:
\begin{equation}
x' = \frac{x - x_{\min}}{x_{\max} - x_{\min}}
\end{equation}
Among them, $x'$ is the normalized value, $x$ is the original value, $x_{\min}$ is the minimum value of the feature, and $x_{\max}$ is the maximum value of the feature.

\subsubsection{Category variable conversion}
In order to be able to use the category variables for cluster analysis, we need to convert them to numeric variables.
\begin{itemize}
    \item \textbf{Personality:} Coded introversion (i) as 0 and extroversion (e) as 1.
    \item \textbf{Student\_Leader:} Already a binary variable (1 is yes, 0 is no).
\end{itemize}

\subsubsection{K-means clustering}
K-means clustering method is used to categorize students into different groups. The algorithm minimizes the Sum of Squared Errors (SSE) within the cluster. SSE is calculated as:
\begin{equation}
SSE = \sum_{i=1}^{k} \sum_{x \in C_i} \|x - \mu_i\|^2
\end{equation}
where $k$ is the number of clusters, $C_i$ is the $i$-th cluster, and $\mu_i$ is the center of the $i$-th cluster. The elbow rule is used to determine the optimal number of clusters $k$ by plotting the SSE curves.

\subsubsection{Principal Component Analysis (PCA)}
In order to reduce the dimensionality of the data and to visualize it, PCA is used in this study. The mathematical principle of PCA involves calculating the covariance matrix:
\begin{equation}
\Sigma = \frac{1}{n} \sum_{i=1}^{n} (x_i - \bar{x})(x_i - \bar{x})^T
\end{equation}
The projection formula into the new space is:
\begin{equation}
Z = XW
\end{equation}
where $X$ is the normalized data matrix, $W$ is the matrix containing the first $k$ eigenvectors, and $Z$ is the projected data matrix.

\subsubsection{Visualization and Cluster Analysis}
After downscaling the data to two principal components by PCA, this study utilizes a 2D scatterplot for visualization. A convex hull of each cluster was also plotted to clearly show the boundary of each cluster.

\section{Results}

\subsection{Data normalization processing}
Through K-means cluster analysis, we identified four main clusters and provided specific career guidance suggestions based on students' academic performance, personality traits, and leadership experiences.

\subsection{K-means cluster analysis}
The K-means clustering algorithm was applied on the normalized data. Specifically, K-means clustering is an unsupervised learning algorithm that iteratively optimizes the distance of data points to the cluster centers so that points within each cluster are as close as possible to each other and points between different clusters are as far away as possible.

Despite the presence of five job categories in the dataset (sales, management, technical, product, and other), we chose K=4 in the K-means clustering.This decision was based on several methodological and practical application considerations:

Statistical and analytical consideration: The main basis for choosing K=4 is to determine the optimal number of clusters through the elbow rule. The elbow rule looks for the inflection point where the curve begins to flatten, the elbow point, by plotting the sum of squared errors (SSE) curves for different values of K. The elbow rule is used to determine the optimal number of clusters for a given cluster. Our analysis shows that the SSE decreases significantly at K = 4 and that the decrease becomes insignificant after this point, suggesting that four clusters best represent the intrinsic structure of the data and minimize the squared error. This choice helps to strike a balance between model complexity and interpretability and avoids overfitting problems.

Homogeneity of clusters and cluster cohesion: By selecting four main clusters, it is possible to ensure a high degree of homogeneity within each cluster, thus maximizing similarities within and differences between clusters. Adding a fifth cluster for the "other" category may introduce heterogeneity within the clusters, thereby reducing the validity of the specific characteristics and actionable recommendations for each cluster. The "other" category contains a wide variety of jobs that may not share common attributes, making it difficult to form a unified cluster.

Practical application and interpretability: From a practical application perspective, the four main clusters (sales, management, technology, and product) are highly aligned with the career paths of the majority of students as well as with the direction of focus of colleges and universities in guiding students. These clusters provide clear and actionable career guidance advice. In contrast, the "Other" category represents a residual group that contains diverse roles that cannot be easily categorized. These roles are more suited to a case-by-case approach rather than guidance through clustering.

Treatment of the "other" category: Although the "other" category was not explicitly clustered, it was recognized and treated separately in the study. For students who fall into the "other" category, individualized career guidance is provided that takes into account their unique characteristics and career interests. This approach ensures that all students receive appropriate support, even if they do not fall into one of the four main clusters. For students in the "other" category, individualized advice is provided through qualitative assessments and one-on-one counseling.

\begin{figure}[htbp]
  \centering
  \includegraphics[width=0.8\textwidth]{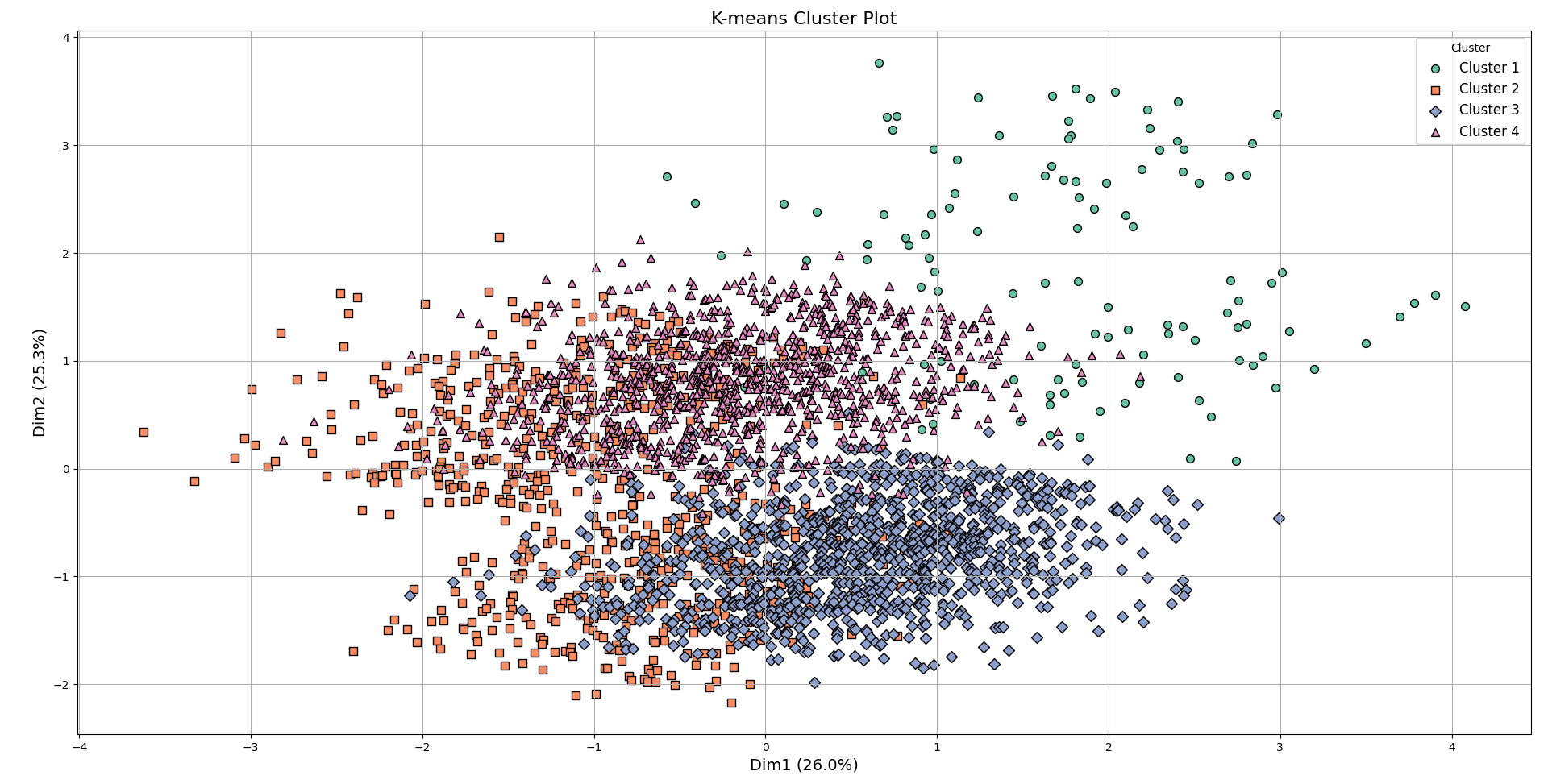}
  \caption{Visualization of K-means clustering results using PCA with convex hulls.}
  \label{fig:pca_clusters}
\end{figure}

To facilitate visualization, Principal Component Analysis (PCA) was used to reduce the dimensionality of the data to two.PCA is a dimensionality reduction technique that projects the data by linear transformation into new coordinate systems whose axes (principal components) are ordered by the magnitude of the variance of the explained data. In this way, the main features of the original data can be presented on a two-dimensional plane, while ensuring that the clustering results are clear and distinct. The visualized image of the clustering results is shown in Figure \ref{fig:pca_clusters}.

An overall view of the cluster plot (Figure \ref{fig:pca_clusters}) shows that the four clusters exhibit significant separation in feature space, which suggests that the groups of students represented by these clusters differ significantly in their characteristics. Students in Cluster 1 (green circular markers) are mainly distributed in the upper right corner of the feature space, indicating high dispersion in certain characteristics; students in Cluster 2 (orange square markers) are more centrally distributed , especially in the left region, reflecting the prominence of specific features; Cluster 3 (blue diamond markers) is more tightly distributed and located in the lower part of the graph, indicating more consistency in multiple features; and Cluster 4 (purple triangle markers) is relatively more evenly distributed and located in the middle of the graph, showing a balance of features.

In the specific analysis of the clustered features, CET4 score is an important indicator of students' English proficiency. In the clustering diagram, students with high CET4 scores are more in Cluster 1, while Cluster 3 and Cluster 4 are relatively balanced in this feature. GPA represents academic achievement, and students with higher GPA are more clearly distributed in Cluster 1 and Cluster 4, showing that these students have a better academic background. Personality traits were determined by the MBTI test, showing a distribution of introverted (i) and extroverted (e) personalities. Extroverted students were more concentrated in Cluster 3 and Cluster 4, and these groups may perform better in teamwork and leadership. In addition, student\_leader experience is more significant in Cluster 1, suggesting that these students have some organizational and leadership experience and are suitable for managerial positions. These analyses provide a scientific basis for students' career guidance and help to formulate personalized career development strategies.

\subsection{Cluster Evaluation and Validation}
Following the K-means clustering analysis, to ensure the accuracy and validity of the clustering results, we conducted an evaluation and validation of the outcomes. The evaluation methods encompassed both internal and external metrics to comprehensively assess the clustering effectiveness.

\subsubsection{Internal Measures}
Internal measures evaluate clustering quality by focusing on the compactness of samples within clusters and the separation between distinct clusters. Common internal measures include the Silhouette Coefficient and the Calinski-Harabasz Index.

\textbf{1. Silhouette Coefficient:} The Silhouette Coefficient measures the difference between the similarity of each point to its cluster members and its similarity to the nearest cluster. It ranges from -1 to 1, with higher values indicating better clustering performance. 

\textbf{2. Calinski-Harabasz Index:} The Calinski-Harabasz Index is the ratio of inter-cluster variance to intra-cluster variance. Higher values indicate better clustering performance.

\subsubsection{External Metrics}
External metrics evaluate the consistency between clustering results and known true labels. Common external metrics include Adjusted Rand Index (ARI) and Homogeneity Score.

\textbf{1. Adjusted Rand Index (ARI):} ARI measures the consistency between actual clustering results and true labels, ranging from -1 to 1. Higher values indicate greater alignment between clustering results and true labels.

\textbf{2. Homogeneity Score:} The Homogeneity Score measures the extent to which each cluster contains samples from only a single category. A higher value indicates greater purity in the clustering results.

\subsection{Clustering feature}
The clustering results identify four distinct groups, each corresponding to a specific employment position. 
The characteristics of the clusters are listed below: 

\vspace{0.5cm}
\noindent\textbf{Cluster 1 (suitable for technical posts)} 

\begin{figure}[htbp]
  \centering
  \includegraphics[width=0.6\textwidth]{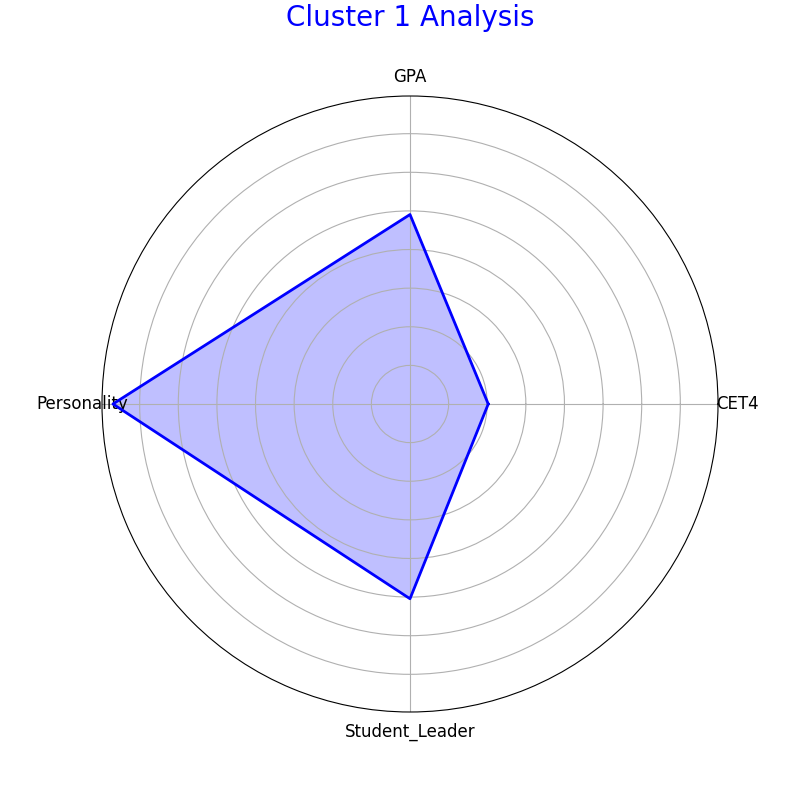}
  \caption{Radar chart analysis for Cluster 1 (Technical Post).}
  \label{fig:radar1}
\end{figure}

\textbf{Radar Chart Analysis}
\begin{itemize}
    \item \textbf{Radar Chart Shape:} As shown in Figure \ref{fig:radar1}, the radar chart shows a significant advantage in the CET6 and GPA dimensions, while other characteristics are more balanced, demonstrating solid academic and language skills. 
\end{itemize}

\textbf{Prevalent Characteristics:}
\begin{itemize}
    \item \textbf{GPA:} above 3.7, reflecting the student's academic ability and drive for self-study. 
    \item \textbf{CET6 score:} higher than 460, showing good language comprehension and application skills. 
    \item \textbf{Personality and student\_leader experience:} introverted, less likely to serve as a student leader, focusing on technical fields. 
\end{itemize}

\textbf{Career Guidance}
\begin{itemize}
    \item \textbf{Recommended position:} technical post 
    \item \textbf{Reason for recommendation:}
    \begin{itemize}
        \item \textit{Combination of theory and practice:} these students have strong theoretical learning ability and self-driven learning motivation, which makes them suitable for technical jobs that require rigorous thinking and complex problem solving ability. 
        \item \textit{Ability to work independently:} introverted traits allow them to excel in work environments that require focus and in-depth analysis, and the ability to concentrate on technical R\&D tasks for long periods of time with fewer external distractions. 
        \item \textit{Technical expertise enhancement:} in conjunction with their academic background, these students can be further encouraged to delve deeper into research and development in specific technical areas, such as software development, data analysis, or engineering design. 
    \end{itemize}
    \item \textbf{Specific recommendations:}
    \begin{itemize}
        \item \textit{Cultivate professional depth:} It is recommended to participate in higher technical training programs and obtain relevant certifications to enhance professional competitiveness. 
        \item \textit{Promote research ability:} Through participation in academic research or technology development projects, students can accumulate practical experience and enhance innovative thinking. 
        \item \textit{Encourage self-reflection:} It is recommended that students maintain their reflective skills during the learning process and regularly assess their personal progress to optimize their learning methods. 
    \end{itemize}
\end{itemize}

\vspace{0.5cm}
\noindent\textbf{Cluster 2 (suitable for management positions)} 

\begin{figure}[htbp]
  \centering
  \includegraphics[width=0.6\textwidth]{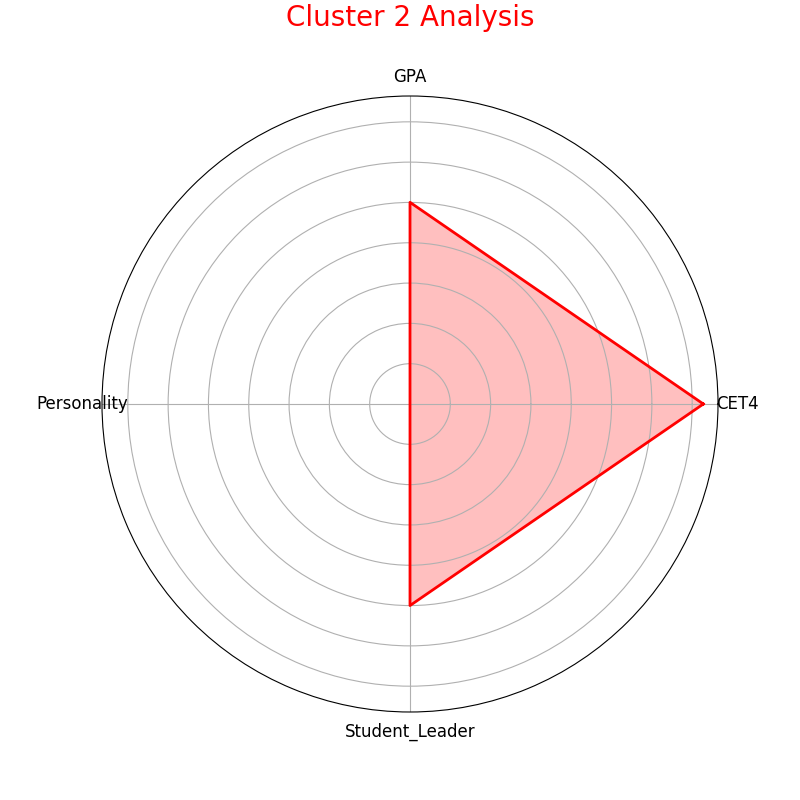}
  \caption{Radar chart analysis for Cluster 2 (Management Post).}
  \label{fig:radar2}
\end{figure}

\textbf{Radar Chart Analysis}
\begin{itemize}
    \item \textbf{Radar Chart Shape:} As shown in Figure \ref{fig:radar2}, the radar chart is balanced and outstanding in terms of GPA, CET6, personality and experience as a student leader, showing comprehensive abilities. 
\end{itemize}

\textbf{General Characteristics:}
\begin{itemize}
    \item \textbf{GPA:} above 3.5 with solid theoretical knowledge. 
    \item \textbf{CET6 score:} higher than 450, strong verbal communication skills. 
    \item \textbf{Personality traits:} outgoing, good at communication. 
    \item \textbf{student\_leader experience:} rich management experience, demonstrating organizational and leadership skills. 
\end{itemize}

\textbf{Career Guidance}
\begin{itemize}
    \item \textbf{Recommended Position:} Management Post 
    \item \textbf{Reason for recommendation:}
    \begin{itemize}
        \item \textit{Comprehensive and coordinating skills:} these students are able to integrate multi-dimensional competencies, quickly adjust strategies in dynamic environments, and are suitable for coordinating and managing roles in complex organizations. 
        \item \textit{Leadership Potential:} experience as a student leader indicates that they already have some leadership skills and team management experience, and can effectively take on management and decision-making roles in the workplace. 
        \item \textit{Strategic vision and decision-making ability:} strong language skills and academic background to provide a clear strategic vision and scientific basis for decision-making in management. 
    \end{itemize}
    \item \textbf{Specific Recommendations:}
    \begin{itemize}
        \item \textit{Leadership development:} participation in leadership courses and internship programs is recommended to enhance organizational coordination and decision-making skills. 
        \item \textit{Teamwork training:} Through team projects and group activities, team awareness and cooperation skills are strengthened. 
        \item \textit{Self-efficacy development:} Encourage students to build positive self-perceptions and enhance confidence in their own abilities to help their career development. 
    \end{itemize}
\end{itemize}

\vspace{0.5cm}
\noindent\textbf{Cluster 3 (suitable for product post)} 

\begin{figure}[htbp]
  \centering
  \includegraphics[width=0.6\textwidth]{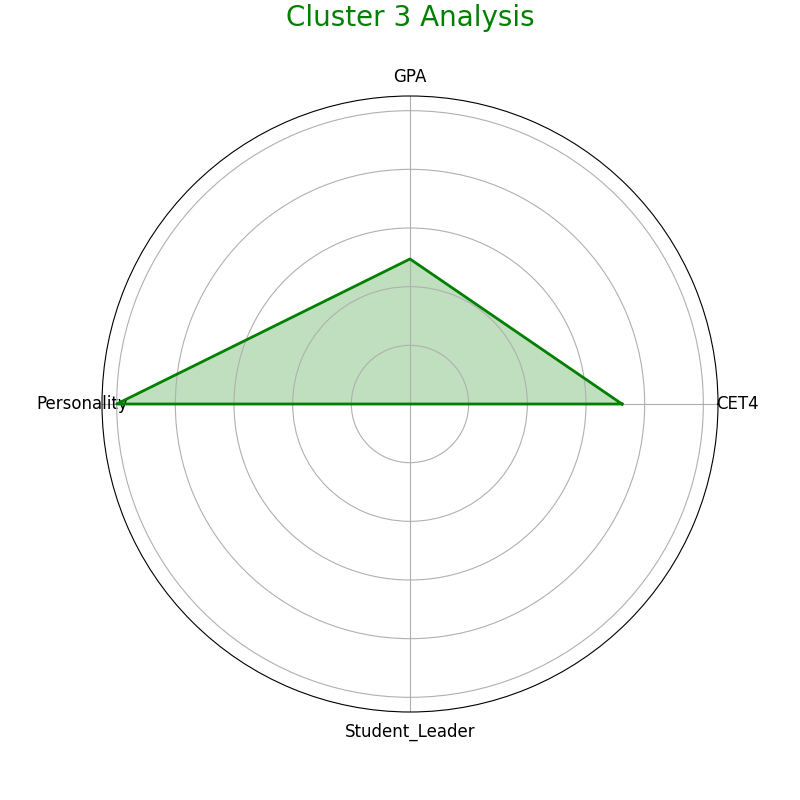}
  \caption{Radar chart analysis for Cluster 3 (Product Post).}
  \label{fig:radar3}
\end{figure}

\textbf{Radar Chart Analysis}
\begin{itemize}
    \item \textbf{Radar Chart Shape:} As shown in Figure \ref{fig:radar3}, the radar chart is outstanding in GPA, CET6 and personality dimensions, and more balanced in other characteristics. 
\end{itemize}

\textbf{Universal Characteristics:}
\begin{itemize}
    \item \textbf{GPA:} above 3.5, solid academic foundation. 
    \item \textbf{CET6 score:} higher than 400, with some language skills. 
    \item \textbf{Personality traits:} outgoing, strong communication and collaboration skills. 
    \item \textbf{Student\_leader experience:} with the ability of organization and coordination. 
\end{itemize}

\textbf{Career Guidance}
\begin{itemize}
    \item \textbf{Recommended Position:} Product Post 
    \item \textbf{Reason for recommendation:}
    \begin{itemize}
        \item \textit{Multi-task management and coordination skills:} product managers need cross-departmental coordination and multi-task management skills, and these students are able to coordinate resources and time effectively. 
        \item \textit{User needs and market analysis:} Their communication skills and outgoing personality enable them to better understand and analyze user needs. 
        \item \textit{Innovation and Change Agents:} the ability to drive innovation and change in product design and development lends itself to working in a fast-paced and rapidly changing market environment. 
    \end{itemize}
    \item \textbf{Specific recommendations:}
    \begin{itemize}
        \item \textit{Creative thinking development:} participation in innovation workshops and product design courses is encouraged to enhance creativity and design thinking. 
        \item \textit{Interdisciplinary learning:} It is recommended to combine technical and marketing knowledge to enhance product development and user experience skills. 
        \item \textit{Practical experience accumulation:} Gain practical experience in product development and market analysis through internships or part-time projects. 
    \end{itemize}
\end{itemize}

\vspace{0.5cm}
\noindent\textbf{Cluster 4 (suitable for sales position)} 

\begin{figure}[htbp]
  \centering
  \includegraphics[width=0.6\textwidth]{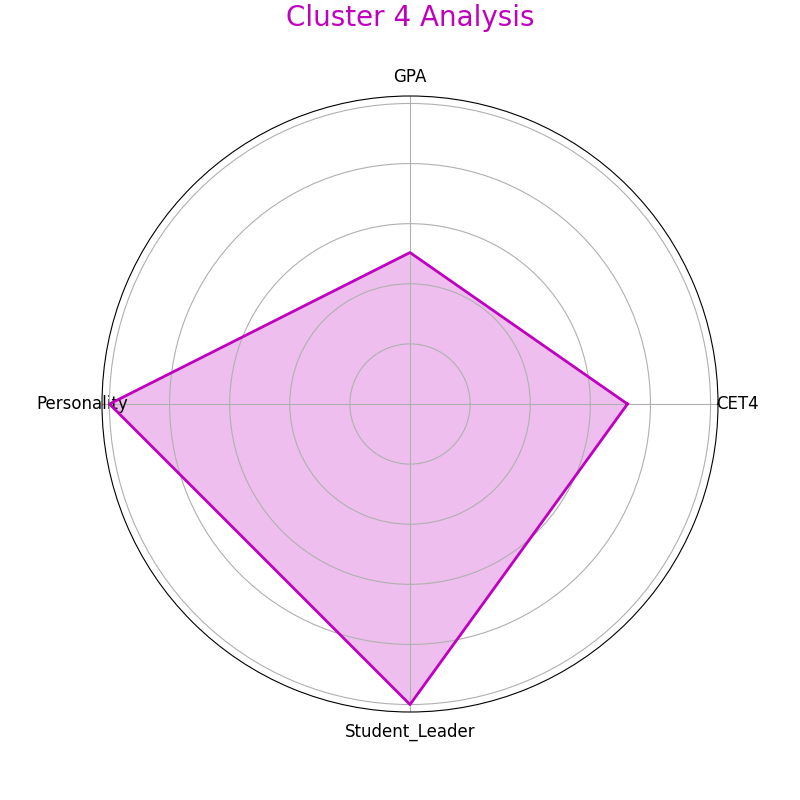}
  \caption{Radar chart analysis for Cluster 4 (Sales Post).}
  \label{fig:radar4}
\end{figure}

\textbf{Radar Chart Analysis}
\begin{itemize}
    \item \textbf{Radar Chart Shape:} As shown in Figure \ref{fig:radar4}, the radar chart is outstanding in personality and CET6 score, other characteristics are more balanced. 
\end{itemize}

\textbf{Universal traits:}
\begin{itemize}
    \item \textbf{CET6 score:} above 400, good verbal communication skills. 
    \item \textbf{Personality traits:} outgoing personality, good at socializing. 
    \item \textbf{Student\_leader experience and GPA:} lower, but with significant strengths in communication and influence. 
\end{itemize}

\textbf{Career Guidance}
\begin{itemize}
    \item \textbf{Recommended Position:} Sales Post 
    \item \textbf{Reason for Recommendation:}
    \begin{itemize}
        \item \textit{Outstanding communication and social skills:} excellent communication and interpersonal skills are key to success in the sales industry, and these students are able to build client relationships and networks quickly. 
        \item \textit{Adaptability \& Stress Resistance:} extroverted students have strong adaptability and stress resistance, and are able to hold up well in a competitive market environment. 
        \item \textit{Persuasion and Negotiation Skills:} Their verbal skills provide the foundation for effective client negotiation and relationship management, enabling them to gain client trust and achieve sales goals during the sales process. 
    \end{itemize}
    \item \textbf{Specific recommendations:}
    \begin{itemize}
        \item \textit{Communication Skills Enhancement:} Participate in elocution and public speaking training to enhance expression and persuasion skills. 
        \item \textit{Customer Relationship Management:} It is recommended to attend customer service courses to learn effective customer management strategies. 
        \item \textit{Practical exercises:} Through internships and sales projects, accumulate practical sales experience and market operation ability. 
    \end{itemize}
\end{itemize}

\subsection{Clustering Evaluation}
The silhouette coefficient is one of the key metrics for evaluating clustering performance. It measures the degree of cohesion within a cluster and the separation between clusters by calculating the silhouette value for each data point. The coefficient ranges from -1 to 1, where a value close to 1 indicates that the data point is tightly grouped within its cluster and well-separated from other clusters, reflecting good clustering quality. The average silhouette coefficient provides an overall assessment of clustering quality, with higher values indicating better clustering outcomes and clearer separation between clusters.

\begin{figure}[htbp]
  \centering
  \includegraphics[width=0.7\textwidth]{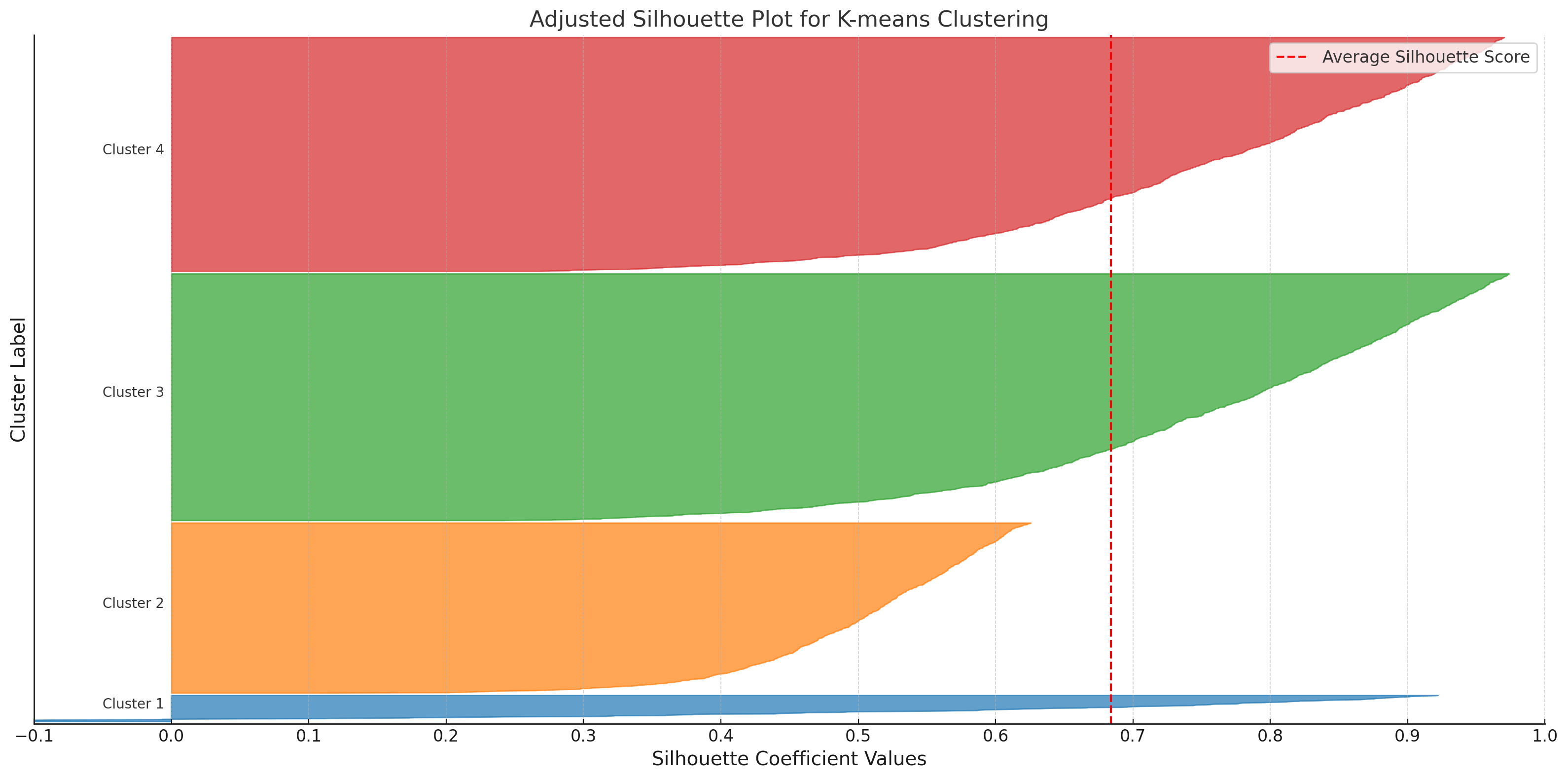}
  \caption{The K-means clustering silhouette plot showing an average silhouette coefficient of 0.684.}
  \label{fig:silhouette}
\end{figure}

As shown in Figure \ref{fig:silhouette}, the K-means clustering silhouette plot shows an average silhouette coefficient of 0.684, suggesting relatively good clustering performance, with most data points tightly distributed within their clusters and well-separated from others. Among the clusters, Cluster 2 exhibits a good silhouette coefficient, with most data points demonstrating high cohesion, indicating effective classification in technical positions. Cluster 3 shows a high silhouette coefficient, reflecting the adaptability of this group of students to management positions, with data points displaying high consistency, indicating strong managerial abilities and good communication skills. Cluster 4 has the highest silhouette coefficient, clearly separating it from other clusters, indicating optimal classification in sales positions. This suggests that students in this cluster may excel in the sales field due to their extroverted personalities and strong communication skills.

\section{Discussion}
The purpose of this study is to explore how to utilize the K-means clustering algorithm to provide career guidance to undergraduate graduates from S University in western China. We analyzed the CET-4 scores, GPAs, personality traits, and student leader experiences of 3000 students and classified them into four main clusters based on these characteristics. Each cluster corresponds to a specific career direction, including technical positions, management positions, product positions，and sales positions. Our main finding is that students with different combinations of characteristics are suitable for different career positions, which provides a scientific basis for personalized career guidance.

These findings suggest that students can be effectively targeted for career guidance by analyzing their academic performance, personality traits and leadership experience. Students with higher academic performance and language skills (GPA higher than 3.7 and CET-4 score higher than 460) are suitable for technical positions because they have advantages in handling complex technical tasks and working independently. Whereas, students with higher academic and language skills and strong leadership and communication skills (GPA higher than 3.5, CET-4 score higher than 450, and have served as student leaders and are outgoing) are suited for managerial positions as they are capable of handling managerial and leadership roles. Students who excel in academic, communication, and leadership skills (GPA above 3.5, CET-4 score above 400, and have served as a student leader with an outgoing personality) are suited for product positions because they are able to multi-task and coordinate. Students with strong communication skills and extroverted personalities (CET-4 scores higher than 400 and extroverted personalities) were suited for sales positions because these characteristics are key strengths in account management and sales. The importance of these findings is that they provide a data-driven framework for colleges and universities to more accurately guide students' career choices and improve their employment success by analyzing and understanding their multifaceted characteristics.

Although we analyzed data from 3000 students, this sample size is still limited relative to larger studies and may affect the representativeness of the results. In addition, this study only considered four main characteristics (CET-4 scores, GPA, personality traits, and experience as a student leader), which fails to encompass all the potential factors affecting career choices. Employment outcomes are also affected by a variety of external factors, such as the economic environment and changes in industry demand, which were not fully taken into account in this study. Based on the above limitations, future research could consider the following: expanding the sample size to improve the representativeness and generalizability of the findings by increasing the sample size; increasing the characteristic variables to introduce more student characteristics, such as internship experience, social activity participation, and personal interest, to provide a more comprehensive analysis; and considering external factors to incorporate external factors, such as the economic environment and industry demand, into the analysis in order to more accurately predict students' career choices and employment success.

\section{Conclusion}
In this study, we analyzed the career guidance of undergraduate graduates from S University in Western China by applying the K-means clustering algorithm. Through detailed processing and analysis of data on CET-4 scores, GPA, personality traits and student cadre experiences of 3000 students, we successfully classified the students into four main clusters and provided targeted career guidance suggestions for each cluster. The results of the study show that students with different combinations of 
characteristics have significant tendencies in career choices, which provides a scientific basis for personalized career guidance and further enhances students' employment success rate.

In this study, K-means clustering algorithm was used to group student data effectively. By assigning students to the most similar groups, the K-means algorithm helped us to identify different groups of students and determine the key features of each group. By minimizing the intra-cluster squared error, the K-means algorithm ensures that students within the same cluster have a high degree of similarity in features, while differences between different clusters are maximized. This process allowed us to make more precise and effective career guidance recommendations for each group's characteristics.

The study found that students with high academic and language skills were better suited for technical positions, while students with strong leadership and communication skills were better suited for management positions. In addition, students with strong academic, communication, and leadership skills were suited for product positions, while students with strong communication skills and extroverted personalities were more suited for sales positions. These findings not only validate existing research on the effectiveness of AI and machine learning in career guidance, but also provide a more nuanced characterization.

Future studies should further expand the sample size to improve the representativeness and generalizability of the findings. At the same time, consideration should be given to introducing more variables of student characteristics, such as internship experience, participation in social activities and personal interests, to provide a more comprehensive analysis. In addition, the inclusion of external factors such as economic environment and industry demand in the analysis will help to predict students' career choices and employment success more accurately.

In summary, this study successfully provided personalized career guidance to students with different combinations of characteristics through the K-means clustering algorithm. This data-driven approach not only significantly improves students' employment success rate, but also provides new ideas and methods for career guidance in universities. Future research directions should include expanding the sample size, increasing the feature variables and considering external factors to further improve the clustering accuracy and guidance effect, so as to better serve the career development needs of students.

\bibliographystyle{unsrt}  
\bibliography{references}  

\end{document}